\documentclass{article}

\usepackage[preprint]{neurips_2024}

\usepackage[utf8]{inputenc} 
\usepackage[T1]{fontenc}    
\usepackage{hyperref}       
\usepackage{url}            
\usepackage{booktabs}       
\usepackage{amsfonts}       
\usepackage{nicefrac}       
\usepackage{microtype}      
\usepackage{xcolor}         
\usepackage{subcaption}
\usepackage{graphicx}
\graphicspath{{figures/}} 
\usepackage{amsmath}
\usepackage{bm}
\usepackage{multicol}
\usepackage{tabularx}
\usepackage{float}
\usepackage[colorinlistoftodos,textsize=footnotesize]{todonotes} 

\providecommand{\sd}[1]{\,{\scriptsize$\pm$#1}}
\providecommand{\best}[1]{{\boldmath\bfseries #1}}

\newcolumntype{Q}{>{\raggedright\arraybackslash}X}

\usepackage[draft]{changes} 

\title{Episodic looking -- using reinforcement learning to bias perception enhances few-shot and continual learning with protection from forgetting}

\author{%
  David Rawlinson \\
  Cerenaut\\
  Melbourne\\
  Australia \\
  \texttt{dave@cerenaut.ai} \\
    \And
   Gideon Kowadlo \\
  Cerenaut\\
  Melbourne\\
  Australia \\
   \texttt{gideon@cerenaut.ai} \\
}

\makeatletter
\def\@noticestring{}
\makeatother

\begin{document}

\maketitle

\begin{abstract}

There is a fundamental contradiction between the desire for models that can generalize, and models that can continuously adapt to recent input with changing statistics. These conditions, captured in few-shot and continual learning settings, can bias or disrupt learned representations. This harms generalization to a broader context, leaving the model unable to overcome input variance.

This paper proposes a solution based on the Complementary Learning System architecture (CLS), in which a Short-Term Memory (STM) delegates generalization to a slow, statistical Long-Term Memory (LTM). Delegation is achieved by working in a space of robust, generalized perceptions produced by the LTM. With this separation, the STM can focus on memorizing recent input and avoiding interference between memories. 

We show that on an image classification task, the resulting CLS model is more resistant to forgetting in a domain-incremental continual learning setting than a baseline ResNet LTM, and generalizes more effectively in a few-shot setting than other continual learners.

In the continual setting, overall accuracy is slightly worse  than existing continual learning models, but with a different mix of performance characteristics. Results analysis suggests that the model is limited by the representations and representational plasticity of the LTM. In future work, we will seek to address this by replaying STM memories to the LTM. 

\end{abstract}

\section{Introduction}

\subsection{Continual learning}
Continual and lifelong learning refer to the idea that a machine learning model should not just be pre-trained once and then used for inference. Instead, it should continue to adapt to the changing statistics of new data (domain-incremental) and novel, unseen classes (class-incremental) \citep{parisi2019continual}. In this paper, we focus on the domain-incremental setting.

In a learned statistical representation, individual samples are not memorized independently but contribute incrementally to latent variables which jointly represent the data. By capturing statistical regularities in the data, the learned representation describes not just the individual samples but the statistics of the population from which those samples were drawn. Attempts to make substantial changes to the representation in response to a few recent samples disrupts the representation of older samples, causing a phenomenon known as ``catastrophic forgetting'' \citep{mccloskey1989catastrophic, goodfellow2013empirical}. Researchers have proposed many methods to minimize interference between memories, which can be broadly categorized as regularization methods (such as Elastic Weight Consolidation \citep{Kirkpatrick2017EWC}), parameter isolation methods \citep{chen2026parameter}, and rehearsal methods, such as experience replay \citep{rolnick2019experience}. Many methods make assumptions about the task, such as awareness of phase changes. A good survey of continual learning techniques and their various assumptions can be found in \citep{wang2024comprehensive}.

A simple form of parameter isolation is to keep a pre-trained encoder frozen and continually learn only a small model which consumes the frozen features. Given effective features, a nearest-class-mean (NCM) classifier on a frozen encoding can be competitive with leading methods \citep{janson2022simple}, and is competitive with the state of the art in some class-incremental learning tasks \citep{zhou2024revisiting}. Recently, a method called RanPAC improved this approach by adding a fixed random projection of the frozen features \citep{mcdonnell2023ranpac}. In this paper, we adopt a similar approach and explore a variety of small continual learning models which consume a set of frozen features in a domain-incremental setting.


\subsection{Complementary Learning Systems}
There are several tensions between the requirements of optimal statistical learning and the requirements of fast, few-shot, and continual learning. Often, the objectives of one type of learning -- such as generalization -- contradicts other objectives, such as rapid adaptation to changing environments, which requires a bias towards more recent input.

The Complementary Learning Systems (CLS) framework provides a biologically inspired solution to achieving both rapid, continual learning of current input and slow, statistical learning of permanent memories. It comprises two memory systems, a Short-Term Memory (STM) and a Long-Term Memory (LTM). In the canonical biologically-inspired formulation, the hippocampus acts as a fast, episodic STM, while the neocortex as LTM gradually accumulates statistical regularities to support slow, stable learning~\citep{mcclelland1995why}. 

More recent work has emphasized how the hippocampus can support rapid inference over episodic experiences, enabling learning without catastrophic forgetting \citep{kumaran2016learning}. Most CLS models focus on episodic STM memorization for replay and consolidation to the LTM, allowing the latter to learn generalized structure \citep{spens2024generative}. 

\subsection{Active Perception in a Complementary Learning System}
Active perception describes a process in which the agent takes actions to influence or improve its own sensory input. These actions might include moving to a new viewpoint, or changing the way sensor data is interpreted and integrated over multiple observations to resolve ambiguity. 

Perception can be both data‑driven and goal‑guided. Bottom‑up perceptual processes build structure from sensory input, whereas top‑down adapts that structure to satisfy task demands, effectively filtering noise and highlighting contextually relevant aspects of the input. Active perception links perception and action in a closed loop: the agent’s current beliefs and goals guide where and how it gathers data, and the resulting data updates its beliefs and subsequent actions. This perspective aligns closely with notions from cognitive neuroscience that perception is not a passive ``snapshot'' of the world but a dynamically constructed interpretation influenced by prior knowledge, expectations, and behavioral goals \citep{bajcsy2018revisiting}.

\begin{figure}[htbp]
     \centering
     \includegraphics[width=0.4\linewidth]{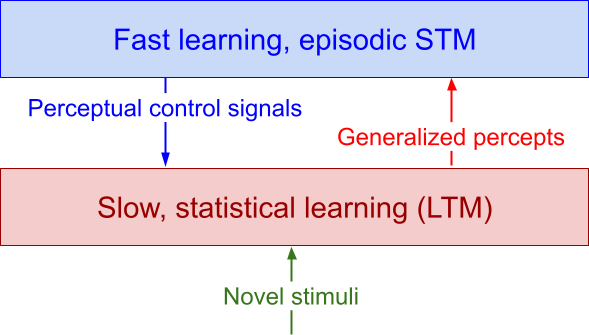}
     \caption{Active perception within a Complementary Learning System could allow a STM to learn rapidly without needing to generalize. In this concept, generalization occurs within the LTM, which is manipulated by the STM to emit stable, familiar representations relevant to the task. Using the generalization ability within the LTM, the STM could even immediately reason about completely novel stimuli, if they are expressed in terms the STM can already understand.}
     \label{fig:active_perception_with_cls}
\end{figure}

In all real environments there are various sources of noise, variance and uncertainty, which can only be overcome by generalization of previous input. Active perception within a CLS might relax some of the tension between slow learning of generalities and fast, few-shot, continual learning by reducing the need for the STM to generalize, while preserving generalization in the system as a whole.

A non-generalizing STM could instead overcome data variance by using active perception to ensure its input is in generalized, familiar terms, even if the external input is novel (figure \ref{fig:active_perception_with_cls}).

In this model, the STM would emit perception control signals which modify the way the LTM represents external input. Since the LTM is a slow, statistical learner, it will recognize and encode novel sensory data as known concepts. For instance, the same image of a bird could be perceived at different levels of specificity — ``a flying animal,'' ``a bird'' or ``a cockatoo'' -- or even in task‑relevant terms such as ``is this a threat?'' 

If an LTM can encode the features of novel stimuli as known representations, an active-perception STM could act immediately (zero-shot) in response to unfamiliar data. Without the requirement to learn generalized representations, it is also easier for the STM learn rapidly and to ensure new memories do not interfere with old ones, because parameters can be safely isolated.


This type of interaction between a fast, episodic memory and a slow, statistical memory has previously been explored in machine learning. In ``Human-inspired Episodic Memory for Infinite Context LLMs''  \citep{fountas2024emllm} augment a frozen large language model with an episodic memory module, retrieving past experiences to construct long-context reasoning without fine-tuning the model. 

Similarly, MemRL~\citep{zhang2026memrl} combines a frozen LLM with an episodic memory. Reinforcement signals are used to update the episodic memory and improve system performance without modifying the LLM. 

Other authors have emphasized the quality and reusability of pre-trained visual features in continual learning tasks \citep{janson2022simple, zhou2024revisiting}. All cited works support the idea that a fast memory system should be able to leverage a slow, statistical model to interpret input in a way that supports rapid learning of context-dependent reasoning.

\section{Method}
This paper describes a specific implementation of the ideas above, applied to the CIFAR-100 image classification dataset \citep{Krizhevsky09learningmultiple} in domain-incremental continual learning and few-shot settings. The experiments are designed to facilitate exploration of the `active perception in CLS' concept. They are not intended to maximize benchmark performance.

We include comparisons to a baseline ResNet-18 model \citep{he2016deep} which represents the conventional statistical learning paradigm, and several CLS STM variants including the FlyModel \citep{shen2023reducing} and SDMLP \citep{bricken2023sparse} continual learners.

Excluding LTM pre-training, all experiments and models learn online. Source code for all methods, models and data are available at:

\url{https://github.com/Cerenaut/episodic_looking}

\subsection{Data}

\begin{figure}[htbp]
     \centering
     \includegraphics[width=0.9\linewidth]{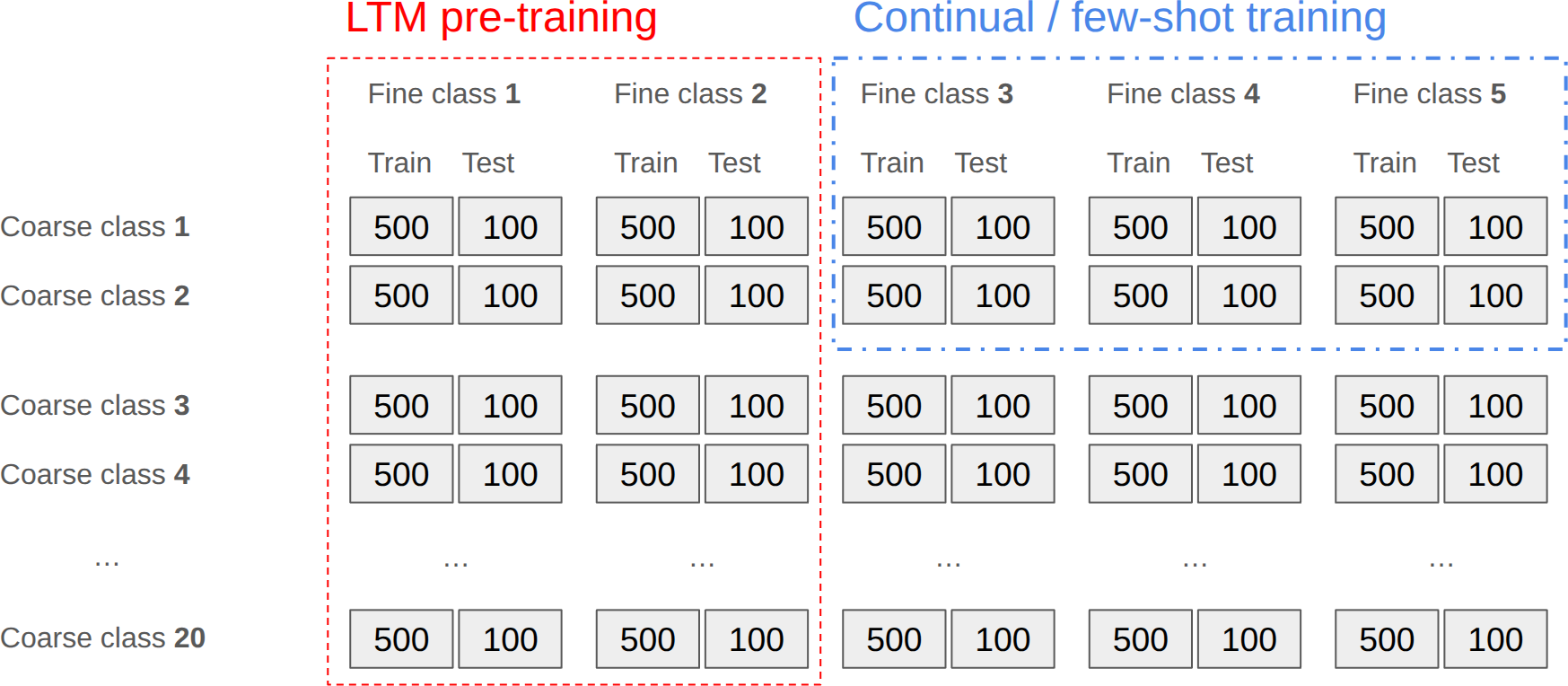}
     \caption{The CIFAR-100 dataset has 20 ``coarse'' classes and 5 ``fine'' sub-classes of each coarse class, yielding a total of 100 distinct classes. The LTM is pre-trained on fine classes 1 and 2 of all coarse classes. Next, individual pairs of coarse classes are selected for continual and few-shot experiments using the unseen fine classes 3, 4 and 5.}
     \label{fig:cifar}
\end{figure}

All reported experiments use the CIFAR-100 image classification dataset \citep{Krizhevsky09learningmultiple}. Each RGB image has size $32\times32\times3$. The dataset has 20 ``coarse'' classes and 5 ``fine'' sub-classes of each coarse class, yielding a total of 100 unique classes (figure \ref{fig:cifar}). The dataset is structured this way to allow evaluation of transfer and generalization between related fine-classes, and discrimination between coarse classes.

To create few-shot and continual learning tasks, we first pre-train the LTM on fine-classes 1 and 2 of all 20 coarse classes. Next, we pick a pair of coarse classes representing ``current'' input and explore few-shot and continual learning of the unseen fine-classes 3, 4 and 5 of the selected coarse pair. Each pair defines a binary classification problem; however, the target class space is fixed to the original 20 coarse classes, with new fine-classes causing input distribution changes compared to pre-training. 

Since this research is an initial feasibility study, we have focused on four pairs of coarse classes which we have labelled A to D (see table \ref{tab:coarse-classes}). These pairs were chosen to (hopefully) represent a range of classification and transfer difficulty levels. 

\begin{table}[htbp]
\caption{Details of the coarse classes used in reported experiments. We formed four pairs (A-D) from these 8 coarse classes.}
\label{tab:coarse-classes}
\centering
\begin{tabular}{llllllll}
    \toprule
   \multicolumn{1}{l}{\textbf{Pair}} &
   \multicolumn{2}{l}{\textbf{Coarse classes}} & \multicolumn{5}{l}{\textbf{Fine classes}} \\
   & & & 1 & 2 & 3 & 4 & 5\\
    \midrule
A & 1 & Aquatic mammals & Beaver & Dolphin & Otter & Seal & Whale \\
A & 2 & Fish & Aquarium fish & Flatfish & Rays & Shark & Trout \\

B &  3 & Flowers & Orchids & Poppies & Roses & Sunflowers & Tulips \\
B &  4 & Food containers & Bottles & Bowls & Cans & Cups & Plates \\

C &  6 & Electrical devices & Clock & Keyboard & Lamp & Telephone & TV \\
C &  7 & Furniture & Bed & Chair & Couch & Table & Wardrobe \\

D &  16 & Reptiles & Crocodile & Dinosaur & Lizard & Snake & Turtle \\
D &  17 & Small mammals & Hamster & Mouse & Rabbit & Shrew & Squirrel \\

\bottomrule
\end{tabular}
\end{table}

\subsection{Model}
\begin{figure}[htbp]
     \centering
     \includegraphics[width=0.8\linewidth]{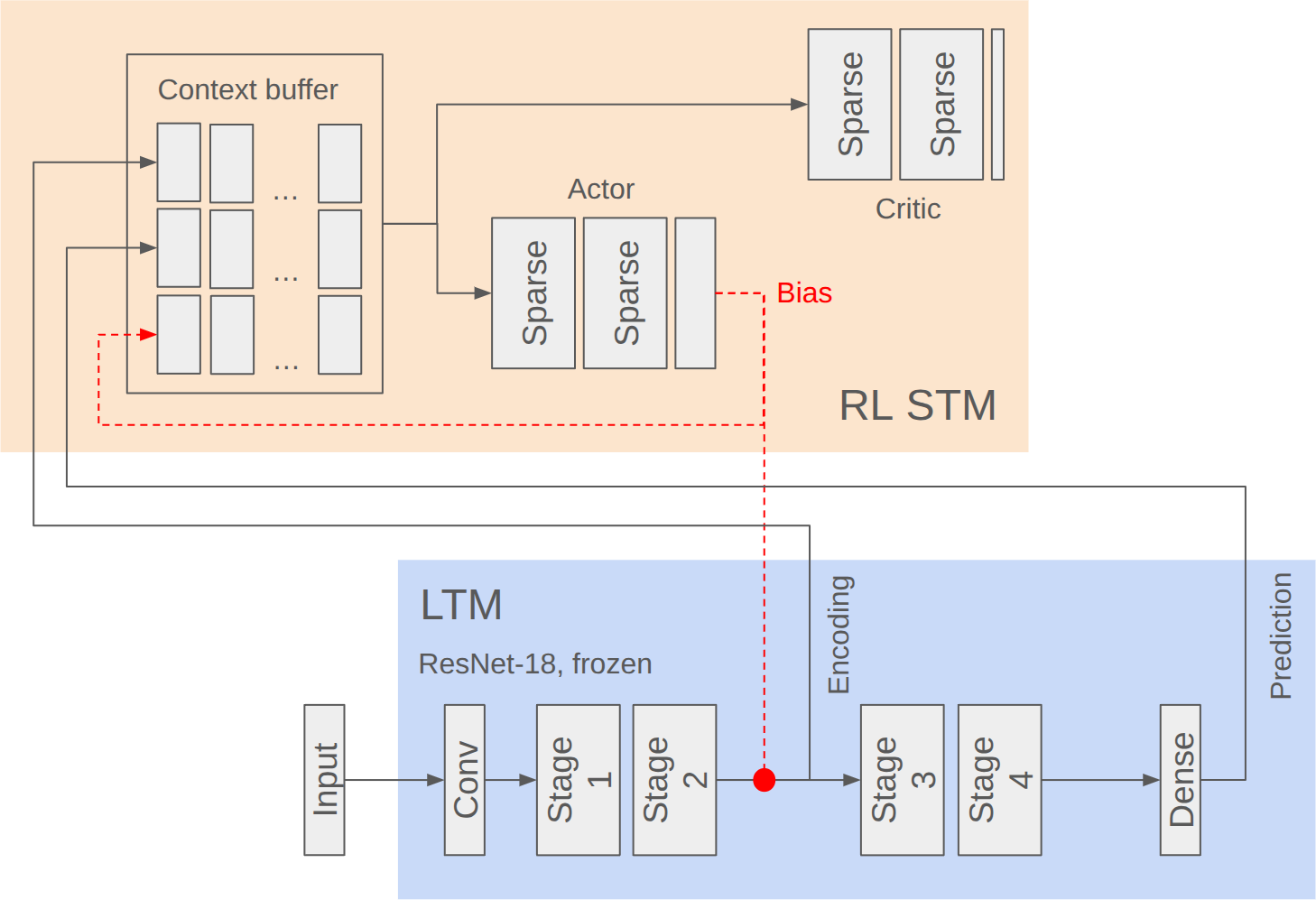}
     \caption{Architecture of the proposed CLS model with a sparse, episodic (RL) Short-Term Memory (STM). The STM is an Advantage Actor-Critic (A2C) model. The actor produces a multiplicative bias output which gates the residual stream of the Long-Term Memory (LTM), a four-stage ResNet-18 model adapted to CIFAR images. STM input includes previous bias output, post-bias LTM encoding and LTM classification logits. These accumulate in a buffer over each episode; the initial encoding is unbiased. The STM is rewarded for producing bias which improves the accuracy of LTM classification.}
     \label{fig:architecture}
\end{figure}

Our proposed model (RL-STM) is a Complementary Learning System (CLS), comprising LTM and STM components (figure \ref{fig:architecture}). The output is always produced by the LTM model, which can be independently pre-trained, fine-tuned and evaluated. This makes analysis of the LTM alone an excellent ablation to characterise the contribution of the STM. The RL-STM can observe the latent states of the LTM and emits a multiplicative bias which gates the forward pass between one pair of residual stages. Unlike most prior instantiations of CLS ( \cite{spens2024generative,sun2023organizing,schapiro2017complementary}), the STM does not augment the LTM, or help it to learn. Instead, the LTM supports the STM, which is the primary actor.

\subsubsection{Long-Term Memory (LTM)}
The LTM is a ResNet-18 architecture \citep{he2016deep} adapted for CIFAR images. The network employs the original post-activation ResNet block, comprising two $3\times3$ convolutions with batch normalization and ReLU activation, with identity shortcuts when feature dimensions are unchanged and learned 1×1 projection shortcuts when spatial resolution or channel dimensionality changes. Further details are provided in Appendix~\ref{app:ltm}.

The LTM was pre-trained for 13 epochs with iid sampling of fine-classes 1 and 2 of all coarse classes. It achieves a test set accuracy of 63.6\% on fine classes 1 and 2 of all coarse classes. The same model achieves a test set accuracy of 31.5\% on fine classes 3, 4 and 5 of all coarse classes. The LTM is deliberately under-trained to retain representational plasticity; one reported ablation uses an LTM with prolonged training (40 epochs).

\subsubsection{Short-Term Memory (RL-STM variant)}
\label{sec:stm}
The RL-STM is implemented as an online, Advantage Actor-Critic (A2C) Reinforcement Learning agent \citep{mnih2016asynchronous}. Supervised learning requires a known, objectively correct output for every input. Reinforcement learning requires only outcome evaluation, which can be sparse and/or delayed, making it a natural choice for episodic interactions with an environment. Although dense and immediate rewards are provided in these experiments, in theory it should be possible to generalize the RL-STM model to more complex environments while retaining the same perceptual architecture.

The RL-STM receives input from the LTM and produces a continuous action output vector which biases the forward pass of the LTM (figure \ref{fig:architecture}). This design is unusual -- most previous work using RL in active perception focuses on where to look (e.g. \citep{chen2021look,mnih2014recurrent}) rather than how to interpret the input. 

Each CIFAR image is observed over an episode of fixed duration $n=8$ steps. Within each step:

\begin{enumerate}
    \item The previous LTM ``encoding'', LTM classification logits and bias are appended to an input buffer. All gradients are discarded.
    \item The RL-STM performs a forward pass, consuming the input buffer, producing a new bias from the actor.
    \item The LTM performs a forward pass, incorporating the bias, producing an output classification. 
    \item A reward is computed for the step.
\end{enumerate}

The bias defaults to 1 at the start of each episode and the LTM performs an initial, unbiased forward pass on the reset observation. Rewards $r_t$ are calculated as the net change in the unit-scaled cross-entropy between predicted class distributions $\mathbf{p}_t$ and target distributions $\mathbf{q}_t$, as defined in Equation~\ref{eqn:reward}.

\begin{equation}
    \begin{aligned}
    R_t &= \exp(-\operatorname{CE}\!\left(\mathbf{q}_t,\mathbf{p}_t\right)),\\
    r_t &= R_t - R_{t-1}, \qquad R_{-1}=0.
    \end{aligned}
    \label{eqn:reward}
\end{equation}

The critic is trained using the Huber loss, as commonly used for value-function estimation in deep reinforcement learning \citep{mnih2016asynchronous}.

The policy is constrained to change gradually using a PPO-style clipped surrogate objective \citep{schulman2017proximal}. At each environment step, the action sampled by the previous policy is re-evaluated under the updated policy, and the resulting probability ratio is clipped to limit the magnitude of the policy change. The clipped objective is weighted by the one-step advantage from the preceding transition.

Both actor and critic comprise 3 fully-connected layers of size 1000, 1000 and (128 or 1) respectively. Model parameters are updated after each environment step with no backpropagation through time or into the LTM. There is no replay buffer. Leaky-ReLU nonlinearities are used in the hidden layers. The model has no regularization features except sparse activation of hidden layers -- details below. To explicitly control learning rate, a Stochastic Gradient Descent optimizer is used with an aggressive learning rate of 0.1 and momentum 0.5.

The RL-STM input buffer contains a concatenation of 3 inputs (figure \ref{fig:architecture}): The previous LTM ``encoding'', taken from the spatially-pooled output of the residual stage that the bias acts on, after the bias is applied, so the RL-STM observes the layer it controls; the LTM classification logits; and the applied bias. These were chosen to help the RL-STM discover how to control the LTM. The unbiased encoding is assumed to contain \emph{potential} interpretations of the input which can be ``selected''. 

The 128-element bias is applied after the second residual stage, which has dimension $C \times H \times W = 128 \times 16 \times 16$. The same bias is applied at all convolutional positions. This choice is constrained by the requirement that the LTM generalize over input variance: bias applied at earlier stages would directly manipulate pixels or relatively low-level image features, for which a learned bias is unlikely to generalize across inputs. RL-STM bias output $\mathbf{b}$ is applied to LTM stage output $\mathbf{y}$ per equation \ref{eqn:bias}.

\begin{equation}
\mathbf{y}' = \mathbf{y} \odot \mathbf{2 \frac{1}{1 + e^{-\gamma\mathbf{b}}}}
    \label{eqn:bias}
\end{equation}

We attempted to apply bias after the final encoding (dimension $C \times H \times W = 512 \times 4 \times 4$) but found that the resulting representation was highly resistant to bias. Even replacing 50\% of the representation elements with random values had no measurable effect on the classification outcome.

These observations suggest a constraint on where top-down perceptual bias can usefully be applied. 
At early processing stages, representations remain too closely tied to the raw sensory input for a learned bias to generalize effectively. At later stages, representations become sufficiently distributed and invariant that substantial perturbation of individual elements has little effect on the resulting classification. 

Intermediate representations therefore occupy a potentially important regime: they contain abstractions sufficiently general to support transfer, while retaining sufficient structure for top-down modulation to influence subsequent perception. This provides a computational motivation for access to intermediate perceptual representations, rather than restricting top-down access to fully abstract, symbolic representations. 

\subsubsection{Sparse activation of STM hidden layers}
\label{sec:sparse}

Sparse representations provide a principled approach to achieving more efficient and robust non-interference between memories \citep{bricken2023sparse,shen2023reducing}. By encoding information such that only a small subset of units is active for a given observation, sparse codes naturally separate features and minimize overlap between representations -- they are highly orthogonal. Sparse representations enable fast learning because updates to one subset of active units minimally affects others, a property first formalized in Kanerva's sparse distributed memory~\citep{kanerva1988sparse}. 

More recently, the Kanerva Machine~\citep{wu2018kanerva} demonstrated that sparse, distributed memory can support generative reconstruction and fast adaptation to new inputs. These findings substantiate the key point that non-interference via sparsity is a powerful enabler of rapid learning, particularly in situations where generalization is not strictly required.

Sparse hidden-layer activity is produced in the RL-STM via a binary mask $\bm{m}$, which is generated via a fixed, random, linear projection $P$ of the input $\bm{x}$ (equation \ref{eqn:project}). $P$ has dimension $[h,i]$ where $h$ is the capacity of the memory and $i$ is the number of elements in the input $\bm{x}$.

A fixed Top-K sparsity is applied to $\bm{p}$ per equation \ref{eqn:topk} yielding the mask $\bm{m}$. $k=32$ was used in our experiments. Finally, the Hadamard (elementwise) product of $\bm{m}$ and the post-activation layer output $\bm{y}$ gives the sparse layer output $\bm{y}'$ (equation \ref{eqn:mask}).

The input and output layers are not sparse. All hidden layers use the same mask; larger layers can be constructed as integer multiples of the base representation size with the mask correspondingly repeated.

\begin{equation}
    \bm{p} = P\bm{x}
    \label{eqn:project}
\end{equation}

\begin{equation}
    \bm{m} = \operatorname{TopK}(\bm{p}, k)
    \label{eqn:topk}
\end{equation}

\begin{equation}
    \bm{y}' = \bm{m} \odot \bm{y}
    \label{eqn:mask}
\end{equation}

To maximize memory utilization, small quantities could be randomly added to values of $P$ which correspond to hidden layer ``cells'' that rarely participate in active cliques. Alternatively, as in ``Continual Backprop'' \citep{dohare2021continual}, disused cells could simply have their parameters in $P$ re-initialized. However, neither option was used in this paper.

\subsection{Comparison and ablation models}
\label{sec:baselines}

\begin{figure}[htbp]
     \centering
     \includegraphics[width=0.8\linewidth]{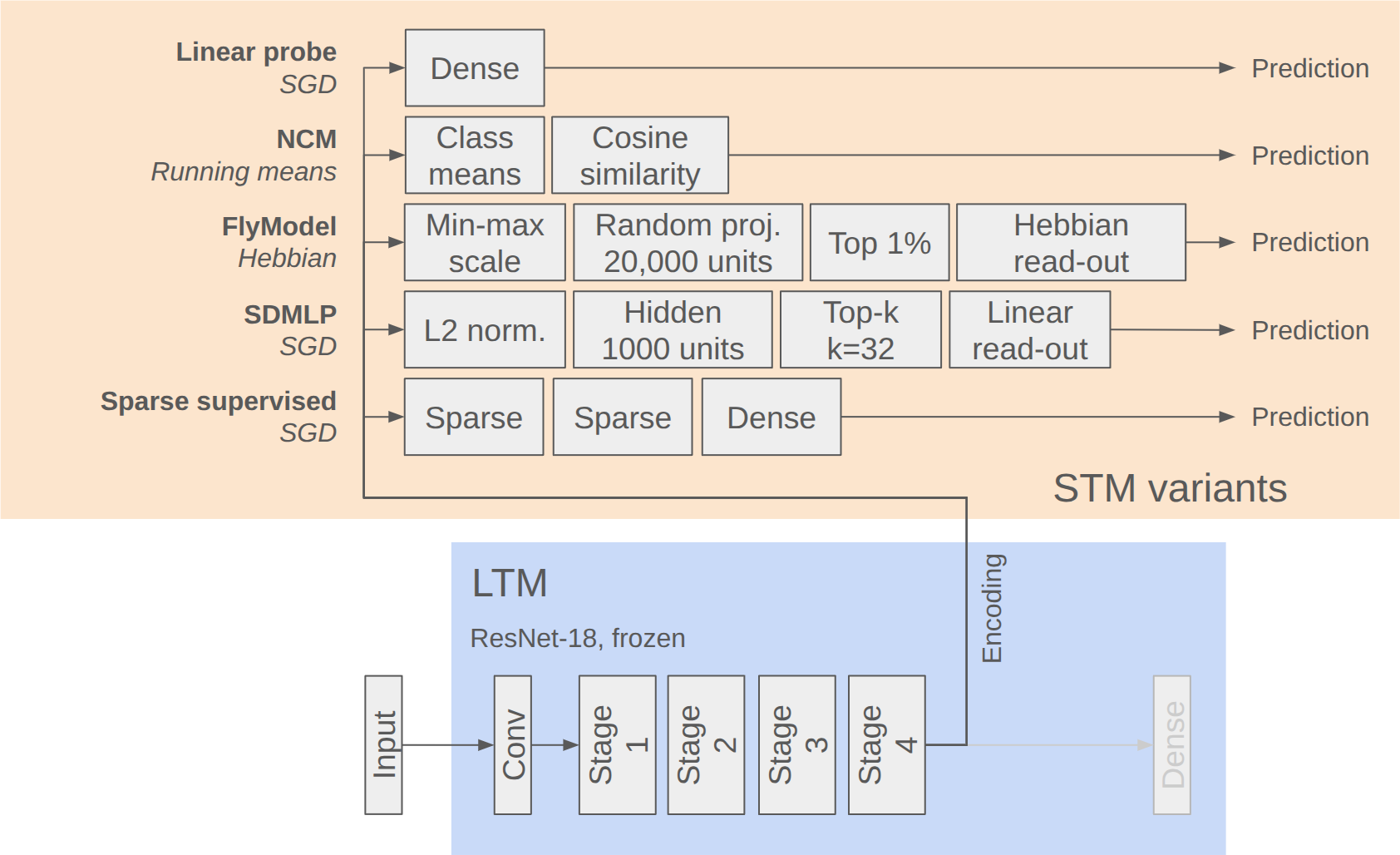}
     \caption{Comparison and ablation STM variants. The RL-STM is compared to some competitive published models (FlyModel, SDMLP) and some simple alternatives such as ablation of various features (linear probe, sparse supervised STM without RL, LTM-only fine-tuning). The Nearest Class Mean (NCM) keeps a running mean of all input samples. The STM variants all consume the 512-dimensional encoding produced by global average pooling of stage 4, which is the input of the LTM's own dense classifier. Only the RL-STM biases the LTM. Italics describe how each STM learns. The linear probe's dense layer is initialized from the LTM's.}
     \label{fig:architecture_other}
\end{figure}

The CLS-inspired RL-STM model combines two ingredients that are known, separately, to reduce interference: a frozen LTM, and hidden layers with sparse activity. To attribute the results to the STM's active-perception loop rather than to these ingredients alone, we compare to STM variants that share the ingredients but classify the LTM's encoding directly, without biasing it (figure \ref{fig:architecture_other}). 

The most relevant prior work is the Sparse Distributed Multi-Layer Perceptron (SDMLP) of \citet{bricken2023sparse}, which learns continually without replay on Split CIFAR-100 by placing a fixed-random-projection and a Top-K hidden layer on top of a frozen, pre-trained convolutional encoder; and its precursor, the FlyModel of \citet{shen2023reducing}, a Hebbian learner derived from the fly olfactory circuit \citep{dasgupta2017neural}. We also include two standard reference heads for continual learning on frozen features: a linear probe (simply a dense layer) and a nearest-class-mean (NCM) classifier. Further implementation details can be found in Appendix \ref{app:other-stm}.

\begin{table}[htbp]
  \caption{The RL-STM could potentially be applied in a wider range of tasks than the other STM variants, because it does not require target labels, a differentiable task loss (the gradient of the loss with respect to its own parameters), or information about phase changes during continual learning. Label-routed updates: the label decides which parameters change, and only the true class's parameters do. Every model we run is phase-blind, but other continual learners such as EWC can exploit this information.}
  \label{tab:assumptions}
  \centering
  \footnotesize
  \begin{tabularx}{\textwidth}{cQXXX}
    \toprule
    
    \textbf{STM variant} & \textbf{Target labels} & \textbf{Label-routed updates} & \textbf{Differentiable task loss} & \textbf{Phase change input} \\
    \midrule
    RL-STM & no & no & no & no \\
    Sparse supervised STM, linear probe, SDMLP & yes & no & yes & no \\
    FlyModel, NCM & yes & yes & no & no \\
    LTM-only fine-tuning & yes & no & yes & no \\
    SDMLP+EWC or EWC (not evaluated) & yes & no & yes & yes \\
    \bottomrule
  \end{tabularx}
\end{table}

The STM variants use information the RL-STM does not assume (table \ref{tab:assumptions}). FlyModel and NCM update only the true class's weights, which reduces interference with memories of other classes. This is label-routed protection, the ``partial freezing'' of \citet{shen2023reducing}, and it requires knowing the correct output, which a reward does not provide. 

\subsection{Experiment design}
\label{sec:experiments}

We performed few-shot and continual learning experiments to illustrate the effects of different learning scenarios on the proposed CLS / STM models and comparison to LTM-only fine-tuning. All STM models are pre-trained on fine-classes 1 and 2 of the selected coarse class pair. Every experiment shares the same pretrained LTM checkpoint. All STM models are trained online, with model updates after each minibatch of images (each episode step of the RL-STM model).

Learning rates and training epoch budgets were optimized for most STM variants, except RL-STM -- see Appendix \ref{app:hparams-opt}. For this purpose a validation set of 100 images per class was extracted from the 500 training images per class, leaving all STM models with 400 training images. 

In the \textbf{continual learning} setting, models are allowed to train on samples from only one unseen fine class during each phase. After completion of the phase, the model is allowed to train on samples from the next unseen fine class. Results are reported for different exposure orders of fine classes 3, 4 and 5. This setting challenges the model to learn from new data while avoiding interference with existing memories. Since the label is always the coarse class, ``learning'' a fine class here means learning to assign its images to the correct coarse class, and ``forgetting'' means losing that ability for fine classes trained in earlier phases.

In the \textbf{few-shot learning} setting, after pre-training, models are allowed to train on different quantities of $N$ samples from one unseen fine class to observe the degree of generalization and transfer that occurs. Generalization refers to performance on the unseen test set of the training class. Transfer refers to learning of class B due to training on another class A, again measured by accuracy on the unseen test set for class B. Results are reported for training on fine classes 3, 4 or 5.

In all settings, accuracy is measured on the test sets of all five fine classes of the selected coarse classes, as a 20-way argmax over all coarse-class logits. However, due to coarse class pairing, every task is in effect a binary classification problem with a 50\% chance level.

\section{Results}
This section compares the RL-STM to the ablation models (Sparse supervised STM, linear probe, NCM and LTM-only fine-tuning) and relevant comparison models from the literature (FlyModel, SDMLP). The models are evaluated in both continual and few-shot learning settings.

\subsection{Continual learning}
\begin{figure}[htbp]
     \centering
     \includegraphics[width=0.8\linewidth]{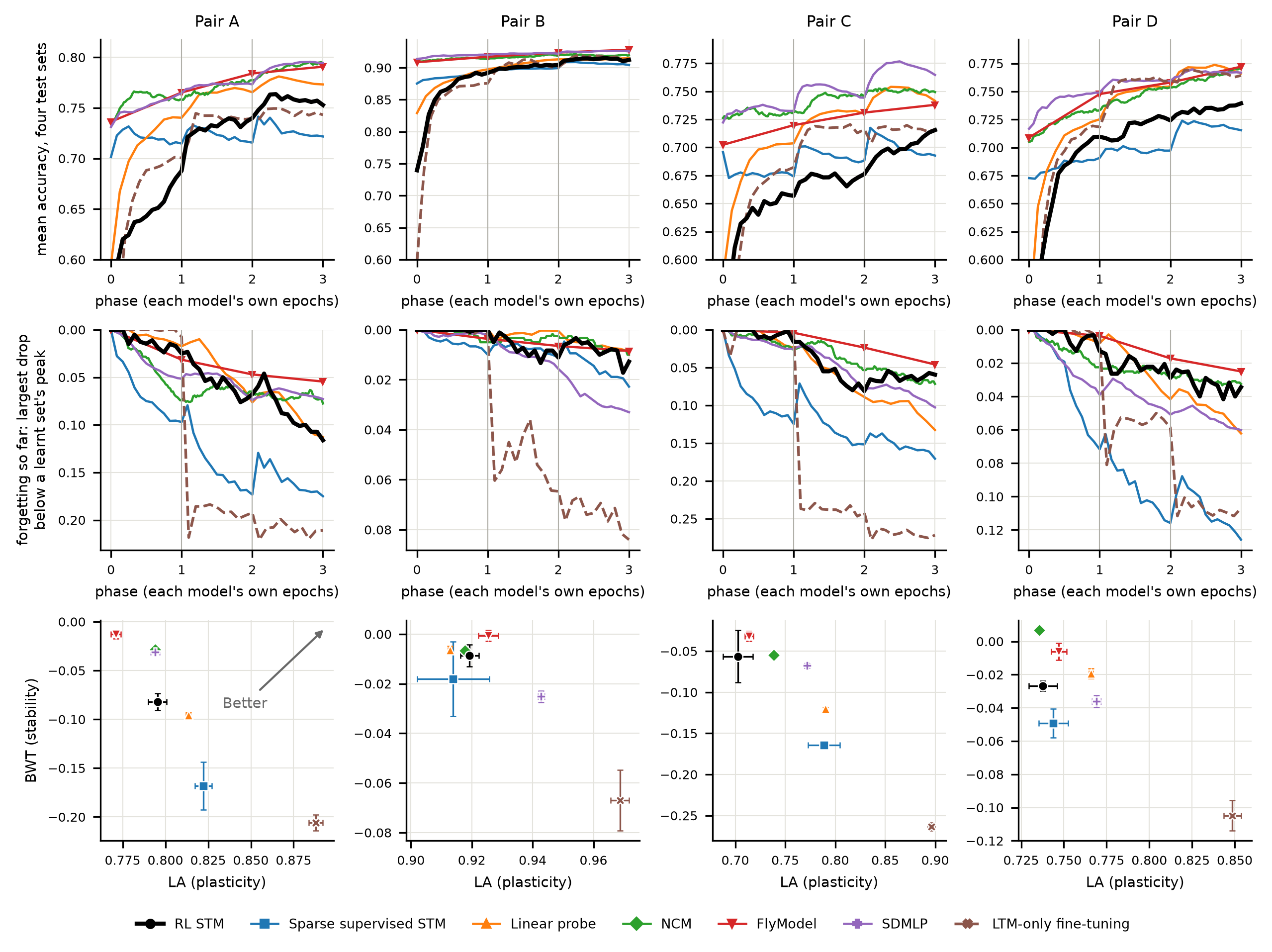}
     \caption{Continual learning results, with test set accuracy for each of the four fine-class sets plotted over the three training phases. Results are averaged over the seeds and orderings of fine classes 3, 4 and 5. Top row: Accuracy over the 4 test sets (1 \& 2, 3, 4, 5). Middle row: Forgetting of any trained fine classes trained so far. Bottom row: Stability / plasticity tradeoff of each model.}
     \label{fig:results-continual}
\end{figure}

Table \ref{tab:results-continual} and figure \ref{fig:results-continual} capture the results for the continual learning setting over the four selected coarse class pairs A-D.  

SDMLP achieves the highest accuracy overall, closely followed by FlyModel and NCM. The RL-STM has overall accuracy close to LTM-only fine-tuning, but their behaviour is actually quite different. 

The bottom row of figure \ref{fig:results-continual} shows the stability-plasticity tradeoff for each model. The RL-STM tends to be close to the FlyModel, SDMLP and NCM with high stability and low plasticity. Consulting the middle row of this figure and table \ref{tab:results-continual} shows that the RL-STM forgets about as much as the SDMLP, but it also learns a bit less overall.

\begin{figure}[htbp]
     \centering
     \includegraphics[width=0.8\linewidth]{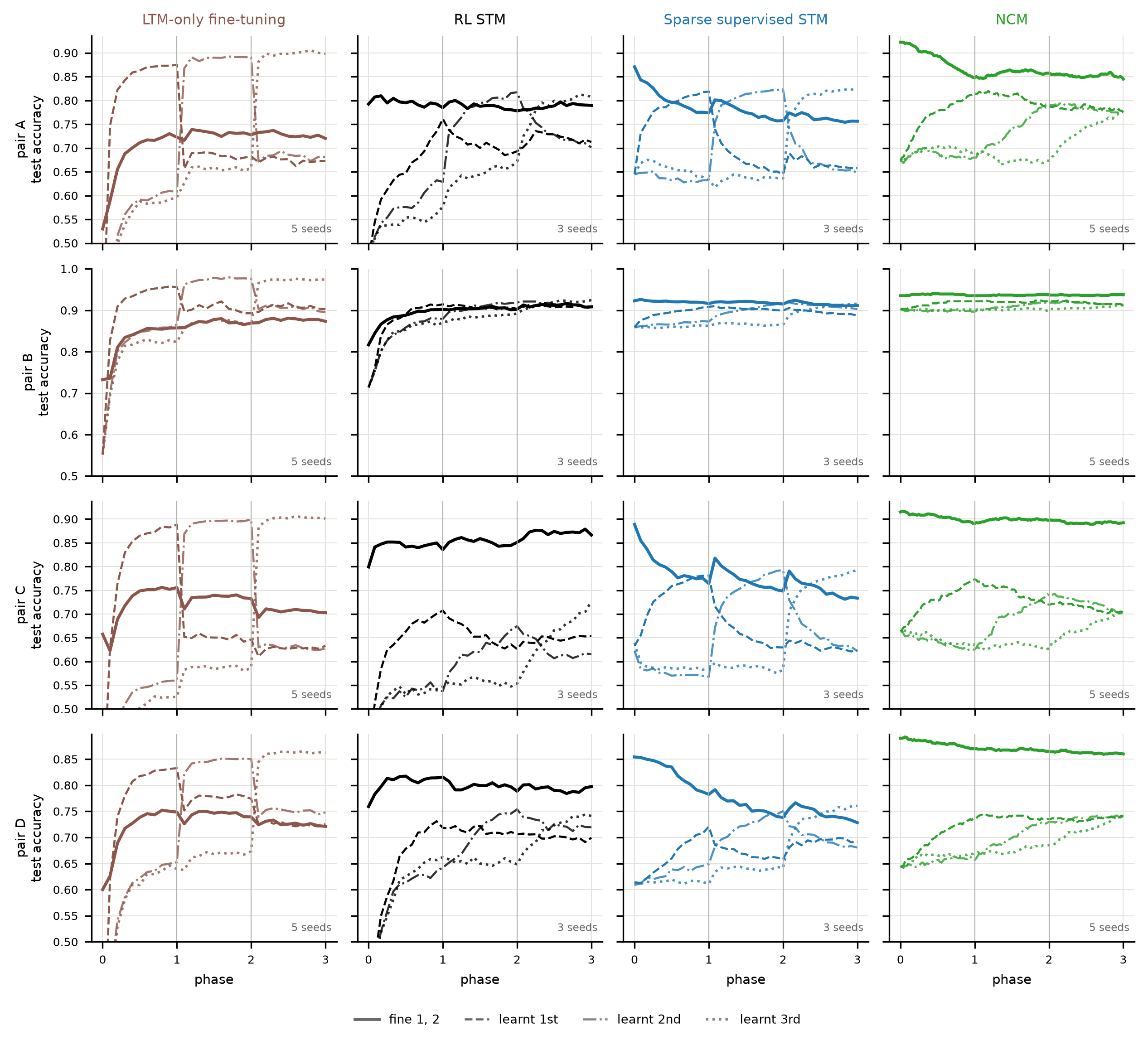}
     \caption{Continual learning results, with test set accuracy for each of the four fine-class sets plotted over the three training phases. Results are averaged over the seeds and orderings of fine classes 3, 4 and 5. The different models shown here learn and forget to different degrees. LTM-only fine-tuning learns the current class more completely than the other models, but forgets a lot. The RL-STM doesn't learn as much, but forgets a lot less. The Sparse supervised STM forgets more than the RL-STM. The NCM ``forgets'' the least, but also learns less than the LTM.}
     \label{fig:results-continual-ablations}
\end{figure}

In contrast, the LTM is far more plastic and far less stable. Figure \ref{fig:results-continual-ablations} explores this aspect by breaking down accuracy over the four fine-class test sets (1 \& 2, 3, 4 and 5). It shows that LTM-only fine-tuning learns the most, but also forgets a lot. As expected, the NCM doesn't really forget anything, but the addition of new training samples changes the classification of some test samples. The RL-STM doesn't learn the new class as much as the LTM, but it forgets a lot less. The sparse supervised STM is effectively the same model as the RL-STM, but trained to classify directly. It learns a little more, but forgets a lot more as well.

The RL-STM, linear probe and LTM-only fine-tuning also demonstrate substantially greater forward transfer (FWT) than all the other models, averaging $+0.19$, $+0.13$ and $+0.34$ respectively versus $-0.006$ to $+0.012$ for the other STM variants (table \ref{tab:results-continual}). This metric assesses how well training on earlier tasks helps the model to learn future tasks.

\begin{table}[htbp]
  \caption{Continual learning test set accuracies over 4 coarse class pairs A-D. Results are averaged over 3 class orderings per pair, and then averaged over pairs. ACC: final average accuracy over the four test sets; LA: learned accuracy on each fine class at the end of its training phase; BWT: change on fine classes 3 to 5 from the end of their phase to the end of the run; BWT$_0$: change on fine classes 1 and 2 over the run; FWT: gain on a class before its own phase; Forgetting: largest drop of a class learnt in the run from its best at a phase boundary. BWT$_0$ and FWT are measured from the end of pre-training, so a model that starts lower scores higher on both. $n$: seeds. ACC is mean $\pm$ one standard deviation over seeds; the standard deviations of the other metrics are at most 0.032 for the RL-STM and the sparse supervised STM, 0.011 for the other heads and 0.012 for LTM-only fine-tuning. Budget: training epochs per phase. Bold: best ACC, LA, BWT and Forgetting in each block.}
  \label{tab:results-continual}
  \centering
  \footnotesize\setlength{\tabcolsep}{1.8pt}
  \begin{tabular}{llrrrrrrrr}
    \toprule
    Pair & Model & $n$ & ACC & LA & BWT & BWT$_0$ & FWT & Forgetting & Budget \\
    \midrule
    A & RL-STM & 3 & 0.753\sd{0.010} & 0.795 & $-0.082$ & $-0.003$ & $+0.182$ & 0.086 & 12 \\
     & Sparse supervised STM & 3 & 0.722\sd{0.012} & 0.822 & $-0.168$ & $-0.114$ & $-0.011$ & 0.169 & 12 \\
     & Linear probe & 5 & 0.773\sd{0.002} & 0.813 & $-0.096$ & $-0.016$ & $+0.158$ & 0.100 & 8 \\
     & NCM & 5 & 0.792\sd{0.000} & 0.794 & $-0.028$ & $-0.078$ & $+0.003$ & 0.035 & 49 \\
     & FlyModel & 5 & 0.791\sd{0.005} & 0.771 & \best{$-0.013$} & $-0.055$ & $+0.029$ & \best{0.017} & 1 \\
     & SDMLP & 5 & \best{0.794}\sd{0.001} & 0.794 & $-0.031$ & $-0.073$ & $+0.027$ & 0.031 & 16 \\
     & LTM-only fine-tuning & 5 & 0.743\sd{0.005} & \best{0.888} & $-0.206$ & $+0.190$ & $+0.346$ & 0.206 & 10 \\
    \addlinespace
    B & RL-STM & 3 & 0.912\sd{0.004} & 0.919 & $-0.009$ & $+0.092$ & $+0.172$ & 0.010 & 12 \\
     & Sparse supervised STM & 3 & 0.904\sd{0.004} & 0.914 & $-0.018$ & $-0.012$ & $+0.009$ & 0.020 & 12 \\
     & Linear probe & 5 & 0.915\sd{0.001} & 0.913 & $-0.006$ & $+0.041$ & $+0.081$ & 0.010 & 8 \\
     & NCM & 5 & 0.919\sd{0.000} & 0.918 & $-0.007$ & $+0.002$ & $-0.006$ & 0.008 & 49 \\
     & FlyModel & 5 & \best{0.928}\sd{0.002} & 0.925 & \best{$-0.001$} & $-0.008$ & $+0.007$ & \best{0.002} & 1 \\
     & SDMLP & 5 & 0.925\sd{0.001} & 0.943 & $-0.025$ & $-0.012$ & $-0.008$ & 0.025 & 16 \\
     & LTM-only fine-tuning & 5 & 0.911\sd{0.008} & \best{0.969} & $-0.067$ & $+0.141$ & $+0.314$ & 0.067 & 10 \\
    \addlinespace
    C & RL-STM & 3 & 0.715\sd{0.009} & 0.703 & $-0.056$ & $+0.067$ & $+0.143$ & 0.058 & 12 \\
     & Sparse supervised STM & 3 & 0.693\sd{0.019} & 0.789 & $-0.164$ & $-0.156$ & $-0.056$ & 0.164 & 12 \\
     & Linear probe & 5 & 0.742\sd{0.000} & 0.790 & $-0.120$ & $-0.003$ & $+0.089$ & 0.120 & 8 \\
     & NCM & 5 & 0.749\sd{0.000} & 0.738 & $-0.055$ & $-0.023$ & $-0.037$ & 0.055 & 49 \\
     & FlyModel & 5 & 0.738\sd{0.002} & 0.714 & \best{$-0.032$} & $-0.030$ & $-0.016$ & \best{0.034} & 1 \\
     & SDMLP & 5 & \best{0.765}\sd{0.002} & 0.772 & $-0.068$ & $-0.036$ & $-0.033$ & 0.068 & 16 \\
     & LTM-only fine-tuning & 5 & 0.716\sd{0.001} & \best{0.896} & $-0.264$ & $+0.045$ & $+0.286$ & 0.264 & 10 \\
    \addlinespace
    D & RL-STM & 3 & 0.739\sd{0.004} & 0.738 & $-0.027$ & $+0.038$ & $+0.263$ & 0.032 & 12 \\
     & Sparse supervised STM & 3 & 0.715\sd{0.008} & 0.744 & $-0.049$ & $-0.126$ & $+0.034$ & 0.051 & 12 \\
     & Linear probe & 5 & 0.766\sd{0.001} & 0.766 & $-0.019$ & $-0.002$ & $+0.216$ & 0.025 & 8 \\
     & NCM & 5 & 0.770\sd{0.000} & 0.736 & \best{$+0.007$} & $-0.030$ & $+0.028$ & \best{$-0.004$} & 49 \\
     & FlyModel & 5 & \best{0.772}\sd{0.005} & 0.747 & $-0.006$ & $-0.025$ & $+0.027$ & 0.007 & 1 \\
     & SDMLP & 5 & 0.767\sd{0.001} & 0.769 & $-0.036$ & $-0.060$ & $+0.027$ & 0.037 & 16 \\
     & LTM-only fine-tuning & 5 & 0.764\sd{0.005} & \best{0.849} & $-0.105$ & $+0.121$ & $+0.445$ & 0.105 & 10 \\
    \midrule
    Mean A--D & RL-STM & 3 & 0.780 & 0.789 & $-0.043$ & $+0.049$ & $+0.190$ & 0.046 &  \\
     & Sparse supervised STM & 3 & 0.759 & 0.817 & $-0.100$ & $-0.102$ & $-0.006$ & 0.101 &  \\
     & Linear probe & 5 & 0.799 & 0.821 & $-0.060$ & $+0.005$ & $+0.136$ & 0.064 &  \\
     & NCM & 5 & 0.808 & 0.796 & $-0.021$ & $-0.032$ & $-0.003$ & 0.023 &  \\
     & FlyModel & 5 & 0.807 & 0.789 & \best{$-0.013$} & $-0.029$ & $+0.012$ & \best{0.015} &  \\
     & SDMLP & 5 & \best{0.813} & 0.819 & $-0.040$ & $-0.045$ & $+0.003$ & 0.040 &  \\
     & LTM-only fine-tuning & 5 & 0.784 & \best{0.900} & $-0.160$ & $+0.124$ & $+0.348$ & 0.160 &  \\
    \bottomrule
  \end{tabular}
\end{table}

\subsection{Few-shot learning}
\label{sec:few-shot}
\begin{figure}[htbp]
     \centering
     \includegraphics[width=\linewidth]{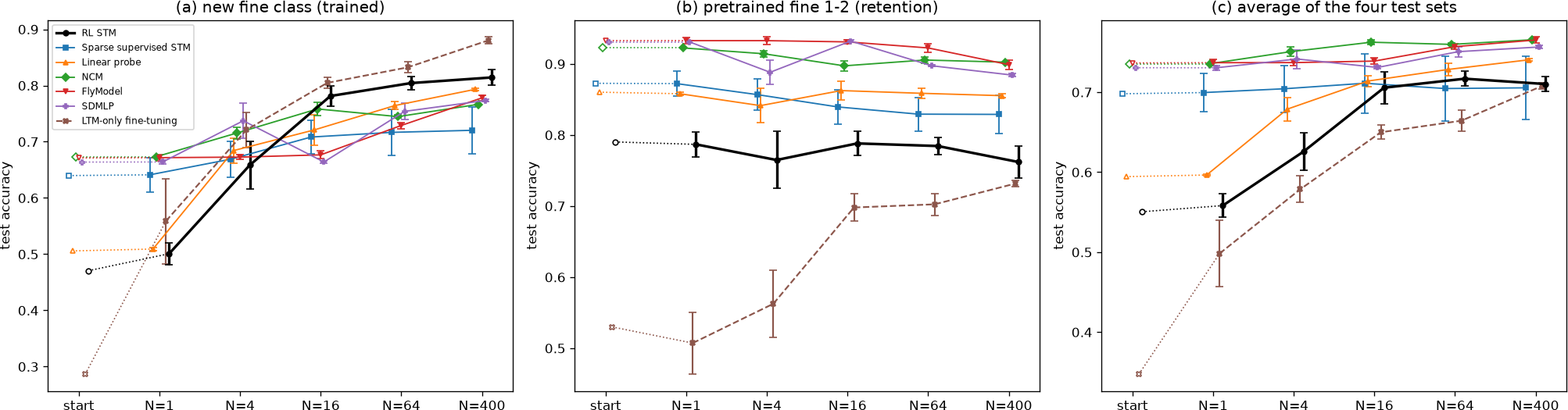}
     \caption{Few-shot results for all STM variants and LTM-only fine-tuning. Results show mean test set accuracy on Pair A, separately trained and evaluated on fine-classes 3, 4, and 5. 10 seeds were used for $N=1$ and $4$ because performance is highly dependent on the images selected. 5 seeds were used for larger $N$. a) Accuracy on the few-shot trained fine-class (mean over 3, 4 and 5). b) Retention of pre-trained fine-classes 1 and 2. c) Mean of the four fine-class sets (1 \& 2, 3, 4, and 5).}
     \label{fig:results-fewshot}
\end{figure}

Table \ref{tab:results-fewshot} and figure \ref{fig:results-fewshot} capture the performance of the STM variants and LTM fine-tuning after few-shot training on Pair A. 

The FlyModel and SDMLP continual learning models do less well in this setting, with lower test accuracy than other models except where $N=1$. They do retain the pre-trained fine classes, which improves their average accuracy.

LTM-only fine-tuning achieves the highest accuracy for $N \ge 16$. Although pre-trained on fine classes 1 and 2, the LTM has not been further trained for a binary classification scenario and therefore performs poorly for small $N$ on the pre-trained fine classes. Increased accuracy for larger $N$ represents transfer to pre-trained fine classes 1 and 2 during training of the few-shot fine class, likely mostly due to elimination of the other coarse classes. 

RL-STM achieves the second highest accuracy on the few-shot trained class for $N \ge 16$ suggesting that it generalizes relatively well. It retains its performance on pre-trained fine classes 1 and 2 albeit from a lower starting point (other configurations of the RL-STM were found to produce higher pre-training accuracy but were not used in this experiment; instead, the configuration was held constant from the continual learning experiment).

The RL-STM struggles with $N = 1$. This may be because the LTM needs almost no bias for optimal classification during STM pre-training. By comparing parameters before and after pre-training, we observe that the Critic experiences substantial change but the Actor doesn't. At the end of pre-training the Actor may have insufficient learning of the policies required to bias the LTM. With only one subsequent image to train on, the Actor might be unable to learn suitable policy for the test images.

\begin{table}[htbp]
  \caption{Few-shot results for all STM variants and LTM-only fine-tuning on Pair A test sets. $N$ denotes the number of training images of each fine-class 3, 4 or 5 presented per coarse class (100 images reserved for validation). All fine-classes were trained separately from checkpoints of the models pre-trained on fine classes 1 and 2. Values are test set accuracies. Top rows: final accuracy on the few-shot trained class. Middle rows: Forgetting on fine classes 1 and 2. Bottom rows: Test set averages over the four fine-classes after few-shot training. Start denotes test accuracy before few-shot training. Training epoch budgets can be found in table~\ref{tab:hparams-opt}.}
  \label{tab:results-fewshot}
  \centering
  \scriptsize\setlength{\tabcolsep}{3pt}
  \begin{tabular}{lrrrrrr}
    \toprule
    Model & Start & $N=1$ & $N=4$ & $N=16$ & $N=64$ & $N=400$ \\
    \midrule
    \multicolumn{7}{l}{\emph{Trained class}} \\
    RL-STM & 0.471 & 0.501\sd{0.018} & 0.658\sd{0.041} & 0.781\sd{0.016} & 0.804\sd{0.010} & 0.814\sd{0.012} \\
    Sparse supervised STM & 0.640 & 0.641\sd{0.029} & 0.669\sd{0.030} & 0.708\sd{0.027} & 0.717\sd{0.036} & 0.720\sd{0.037} \\
    Linear probe & 0.506 & 0.509\sd{0.002} & 0.684\sd{0.021} & 0.721\sd{0.024} & 0.764\sd{0.006} & 0.793\sd{0.001} \\
    NCM & 0.673 & \best{0.673}\sd{0.000} & 0.716\sd{0.010} & 0.758\sd{0.011} & 0.745\sd{0.003} & 0.767\sd{0.000} \\
    FlyModel & 0.671 & 0.672\sd{0.005} & 0.673\sd{0.004} & 0.677\sd{0.006} & 0.729\sd{0.005} & 0.778\sd{0.005} \\
    SDMLP & 0.664 & 0.664\sd{0.004} & \best{0.737}\sd{0.029} & 0.664\sd{0.002} & 0.754\sd{0.013} & 0.773\sd{0.002} \\
    LTM-only fine-tuning & 0.287 & 0.559\sd{0.072} & 0.721\sd{0.030} & \best{0.805}\sd{0.009} & \best{0.832}\sd{0.009} & \best{0.880}\sd{0.005} \\
    \addlinespace
    \multicolumn{7}{l}{\emph{Forgetting, fine classes 1 and 2}} \\
    RL-STM & -- & 0.012\sd{0.019} & 0.040\sd{0.031} & 0.014\sd{0.011} & 0.012\sd{0.011} & 0.026\sd{0.016} \\
    Sparse supervised STM & -- & 0.001\sd{0.002} & 0.017\sd{0.020} & 0.029\sd{0.016} & 0.041\sd{0.013} & 0.042\sd{0.019} \\
    Linear probe & -- & 0.003\sd{0.002} & 0.028\sd{0.018} & 0.010\sd{0.009} & 0.017\sd{0.005} & 0.022\sd{0.002} \\
    NCM & -- & \best{0.000}\sd{0.000} & 0.010\sd{0.004} & 0.026\sd{0.007} & 0.020\sd{0.004} & 0.023\sd{0.000} \\
    FlyModel & -- & \best{0.000}\sd{0.000} & \best{0.001}\sd{0.001} & 0.001\sd{0.001} & 0.009\sd{0.005} & 0.032\sd{0.008} \\
    SDMLP & -- & \best{0.000}\sd{0.000} & 0.043\sd{0.016} & \best{0.000}\sd{0.000} & 0.035\sd{0.002} & 0.048\sd{0.004} \\
    LTM-only fine-tuning & -- & 0.049\sd{0.026} & 0.021\sd{0.022} & \best{0.000}\sd{0.000} & \best{0.000}\sd{0.000} & \best{0.000}\sd{0.000} \\
    \addlinespace
    \multicolumn{7}{l}{\emph{Average of the four test sets}} \\
    RL-STM & 0.551 & 0.558\sd{0.014} & 0.626\sd{0.022} & 0.706\sd{0.018} & 0.717\sd{0.008} & 0.710\sd{0.008} \\
    Sparse supervised STM & 0.698 & 0.700\sd{0.022} & 0.704\sd{0.028} & 0.711\sd{0.033} & 0.705\sd{0.036} & 0.706\sd{0.036} \\
    Linear probe & 0.594 & 0.596\sd{0.001} & 0.679\sd{0.014} & 0.714\sd{0.006} & 0.729\sd{0.007} & 0.741\sd{0.001} \\
    NCM & 0.736 & 0.736\sd{0.000} & \best{0.751}\sd{0.005} & \best{0.762}\sd{0.003} & \best{0.760}\sd{0.001} & \best{0.766}\sd{0.000} \\
    FlyModel & 0.737 & \best{0.737}\sd{0.004} & 0.737\sd{0.003} & 0.739\sd{0.004} & 0.757\sd{0.003} & \best{0.765}\sd{0.003} \\
    SDMLP & 0.731 & 0.731\sd{0.003} & 0.742\sd{0.011} & 0.731\sd{0.002} & 0.751\sd{0.006} & 0.756\sd{0.002} \\
    LTM-only fine-tuning & 0.348 & 0.498\sd{0.039} & 0.579\sd{0.016} & 0.650\sd{0.008} & 0.664\sd{0.011} & 0.708\sd{0.003} \\
    \bottomrule
  \end{tabular}
\end{table}

\subsection{Episodic dynamics}
\label{sec:episodic-dynamics}
\begin{figure}[htbp]
     \centering
     \includegraphics[width=0.6\linewidth]{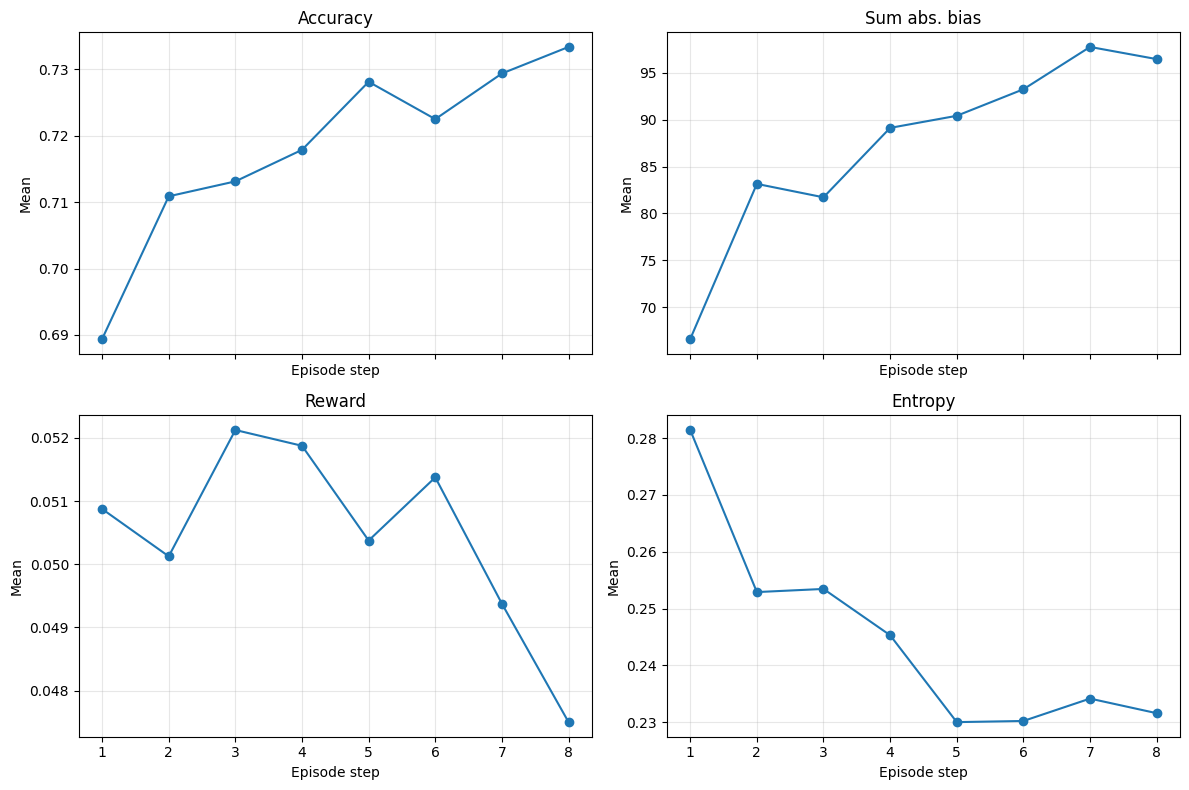}
     \caption{Analysis of STM behaviour within an episode. These data were collected from processing 500 images of Pair A fine-class 3 on which the RL-STM model was already trained. Accuracy and entropy are derived from the LTM output logits.}
     \label{fig:intra-episode}
\end{figure}

To explore the interaction of RL-STM and LTM, figure \ref{fig:intra-episode} depicts key quantities which change during each episode, sampled from processing of 500 images. The majority of accuracy and bias magnitude is obtained from the first forward pass, with a further gain of approximately 5\% accuracy within the arbitrary 8-step episode time limit. These results confirm that the model is not simply emitting a constant learned bias in the first forward pass, but adapting the bias to refine LTM output. 

It is possible that this adaptation is also crucial during the training process, but this has not been further explored.

Logit entropy is consistently low despite changes in classification outcome (increasing accuracy shows that the outcome must change), indicating that the LTM is typically confident in its changing predictions.

How does the bias behave? The original concept was that the STM should induce the LTM to try to interpret the input in a particular way, searching for a good response. Since the classification does change within an episode, this is occurring to some extent. 

To analyze this further, we produced prototype bias vectors for each coarse $\times$ fine class combination as the mean of unit-length final bias vectors over all training images from Pair A, after a continual learning task. Predictions were generated for each test image bias vector by finding the closest class prototype, using cosine distance.

Analysis showed this predicted the coarse class with 76\% accuracy but only predicted the fine-class with 27\% accuracy, against a 10\% chance level (5 fine classes $\times$ 2 coarse classes). This suggests the STM may be overriding the LTM output rather than merely influencing it. However, it is possible that the STM induces a ``Whale'' classification by insisting that it looks like a ``Dolphin'', rather than by forcing a Whale to be a Whale. The RL-STM never has access to fine-class labels, not even via rewards, so they only exist as statistical regularities in the data.

The trained bias on average moves LTM features slightly away from their own fine-class prototypes (cosine distance 0.81 to 0.78). It is possible they move towards a superposition of prototypes, or towards a prototype which more firmly defines the coarse class.

\subsubsection{Prolonged LTM pre-training}
\label{sec:prolonged-ltm-pre-training}
We also performed some tests of the RL-STM architecture using a LTM checkpoint with prolonged pre-training (40 epochs). While these results were preliminary, RL-STM performance was overall significantly worse, suggesting that the \emph{representational} plasticity of the LTM is important for the combined architecture. Representational plasticity refers to the ability for latent representations to be transformed from one prototype to another via bias, rather than capacity for further training. Explicit promotion of representational plasticity -- possibly via label noise injection -- might significantly enhance the combined architecture.

\section{Discussion and Conclusion}
To conclude, we will revisit the original objective of this research: to explore the capabilities of an active-perception STM in a CLS architecture, and to discover whether it can learn to modulate the behaviour of a slow-learning LTM to maximize a reward.

\textbf{Is it possible for a RL-STM to learn to modulate the output of a LTM to achieve a task?} Yes. This is a relatively challenging architectural choice because the RL-STM has a large output dimension, feedback via rewards is less specific, and it can only modify the classification output indirectly. In addition, all models were trained online. Despite these constraints, overall test set accuracy was comparable to the other approaches such as LTM fine-tuning in both continual and few-shot settings. 

As described in table \ref{tab:assumptions}, the RL-STM can be used in a wider variety of task settings because it does not require target labels, a differentiable task loss, label-routed updates or phase change visibility, all of which are required by one or more comparison models.

\textbf{Does the RL-STM architecture demonstrate effective generalization?} Yes. In fact, in the few-shot setting where N is 16 or more, the RL-STM is more accurate than all other models tested, except LTM fine-tuning. The RL-STM also achieved higher foward transfer in the continual setting than all other models except LTM fine-tuning. This is despite having no features to promote regularization and generalization other than sparse activation and the episodic loop. 

\textbf{Was there evidence that the ``episodic looking'' concept could enhance model performance?} Yes. 
Our baseline is LTM fine-tuning. We expected that freezing the LTM and adding sparse activation to the STM would provide some degree of protection from forgetting in a continual setting, and this was confirmed by experiment; all STM variants evidenced less forgetting than LTM fine-tuning.

We expected that an episodic forward process would explore a wider range of observations than a single forward pass, resulting in less forgetting in a continual setting and improved generalization in a few-shot setting (both due to different expressions of the same input). The RL-STM performed well in the few-shot setting and was resistant to forgetting in the continual setting. It demonstrated substantially higher forward transfer than all other STM variants. It outperformed the sparse supervised STM, which is basically the same model with a supervised loss instead of the TD-0 target (one-step return).

Analysis of accuracy changes within an episode showed that while most of the answer is obtained in the first forward pass, subsequent forward passes improve the resulting classification by approximately 5\%. This requires an episodic or recurrent framing. The few-shot results suggest that all forward passes may contribute to training and may be responsible for a larger fraction of model performance than the in-episode uplift. To confirm this, an RL-STM with only one step per episode could be evaluated.

\subsection{Limitations}
The results reported here are only part of just one benchmark dataset, but they do characterise the performance of the RL-STM architecture and motivate further research. Overall, the RL-STM presents an interesting set of tradeoffs which differentiate it from the comparison and ablation models. However, it was not possible to beat the FlyModel and SDMLP STM variants in the continual setting. 

During development of the RL-STM we explored methods to improve continual learning performance by conditioning the input and enhancing trainable parameter initialization. These methods were successful in improving cell utilization to over 95\% but did not improve final test set accuracy in continual learning.

Further investigation showed that the limiting factor is the frozen LTM representation. In Pair A (aquatic mammals vs fish), almost all the forgetting is due to re-labelling sharks (fine-class 4) to dolphins (fine class 2) and whales (fine-class 5) or vice-versa. These animals do look alike and we found the sparse masks of sharks and whales overlap by an average of 0.82, almost as much overlap as two images of the same class (0.9).

These small differences \emph{can} be exploited in a supervised classifier (the SDMLP achieved 81.3\% overall in continual learning, +3.3\%  over the RL-STM), but the RL-STM must also induce the frozen LTM classifier to distinguish them, and they may be bound together in later residual stages. We found that the LTM was highly resistant to bias after stage 4 and after prolonged pre-training. We conclude that the combined architecture is sensitive to the LTM's representational plasticity.

\subsection{Future work}
To enhance the overall performance of the RL-STM we conclude that the LTM must be updated -- this is the role of consolidation in the CLS architecture. To achieve this, a replay module is required to generate samples for LTM learning. 

To test this architecture, we propose to add another continual learning experimental condition in which the samples are presented in a single correlated stream. This setting is designed to test the limits of LTM-only fine-tuning, which is expected to be dependent on iid sampling. Model variants which replay randomly sampled data to the LTM should be necessary for optimal learning. This setting is designed to mimic the conditions experienced by a single mobile robot which must learn a novel environment.

Given a single correlated stream, the replay module would face many of the same challenges encountered by the current STM variants, such as continual learning without forgetting earlier input, and few-shot generalization from incremental observation of input samples. 



\bibliographystyle{abbrvnat}
\bibliography{bibliography}

@inproceedings{he2016deep,
  title={Deep Residual Learning for Image Recognition},
  author={He, Kaiming and Zhang, Xiangyu and Ren, Shaoqing and Sun, Jian},
  booktitle={Proceedings of the IEEE Conference on Computer Vision and Pattern Recognition},
  pages={770--778},
  year={2016}
}

@techreport{Krizhevsky09learningmultiple,
  author      = {Alex Krizhevsky},
  title       = {Learning multiple layers of features from tiny images},
  institution = {University of Toronto},
  year        = {2009}
}

@article{parisi2019continual,
  title={Continual lifelong learning with neural networks: A review},
  author={Parisi, German I and Kemker, Ronald and Part, Jose L and Kanan, Christopher and Wermter, Stefan},
  journal={Neural networks},
  volume={113},
  pages={54--71},
  year={2019},
  publisher={Elsevier}
}

@inproceedings{chen2026parameter,
  title={Parameter-efficient Continual Learning for Enhancing Plasticity without Forgetting under Limited Model Capacity},
  author={Chen, Yitian and Zhang, Shigeng and Liu, Xuan and Lu, Mingming and Chen, Kai and Zhu, Hongye and Chen, Xinning},
  booktitle={Proceedings of the IEEE/CVF Conference on Computer Vision and Pattern Recognition},
  pages={10789--10798},
  year={2026}
}

@incollection{mccloskey1989catastrophic,
  title     = {Catastrophic Interference in Connectionist Networks: The Sequential Learning Problem},
  author    = {McCloskey, Michael and Cohen, Neal J.},
  booktitle = {Psychology of Learning and Motivation},
  volume    = {24},
  pages     = {109--165},
  year      = {1989},
  publisher = {Academic Press},
  doi       = {10.1016/S0079-7421(08)60536-8}
}

@article{goodfellow2013empirical,
  title   = {An Empirical Investigation of Catastrophic Forgetting in Gradient-Based Neural Networks},
  author  = {Goodfellow, Ian J. and Mirza, Mehdi and Xiao, Da and Courville, Aaron and Bengio, Yoshua},
  journal = {arXiv preprint arXiv:1312.6211},
  year    = {2013},
  url     = {https://arxiv.org/abs/1312.6211}
}

@article{mcclelland1995why,
  title     = {Why there are complementary learning systems in the hippocampus and neocortex:
               Insights from the successes and failures of connectionist models of learning and memory},
  author    = {McClelland, James L. and McNaughton, Bruce L. and O'Reilly, Randall C.},
  journal   = {Psychological Review},
  volume    = {102},
  number    = {3},
  pages     = {419--457},
  year      = {1995},
  doi       = {10.1037/0033-295X.102.3.419}
}

@article{kumaran2016learning,
  title     = {What Learning Systems do Intelligent Agents Need? {Complementary} Learning Systems Theory Updated},
  author    = {Kumaran, Dharshan and Hassabis, Demis and McClelland, James L.},
  journal   = {Trends in Cognitive Sciences},
  volume    = {20},
  number    = {7},
  pages     = {512--534},
  year      = {2016},
  doi       = {10.1016/j.tics.2016.05.004},
  url       = {https://www.sciencedirect.com/science/article/pii/S1364661316300432}
}

@article{spens2024generative,
  title     = {A generative model of memory construction and consolidation},
  author    = {Spens, Eleanor and Burgess, Neil},
  journal   = {Nature Human Behaviour},
  volume    = {8},
  pages     = {225--240},
  year      = {2024},
  doi       = {10.1038/s41562-023-01799-z},
  url       = {https://www.nature.com/articles/s41562-023-01799-z}
}

@article{sun2023organizing,
  title     = {Organizing memories for generalization in complementary learning systems},
  author    = {Sun, Weinan and Advani, Madhu and Spruston, Nelson and Saxe, Andrew and Fitzgerald, James E.},
  journal   = {Nature Neuroscience},
  volume    = {26},
  pages     = {1438--1448},
  year      = {2023},
  doi       = {10.1038/s41593-023-01382-9}
}

@article{schapiro2017complementary,
  title     = {Complementary learning systems within the hippocampus: a neural network modelling approach
               to reconciling episodic memory with statistical learning},
  author    = {Schapiro, Anna C. and Turk-Browne, Nicholas B. and Botvinick, Matthew M. and Norman, Kenneth A.},
  journal   = {Philosophical Transactions of the Royal Society B: Biological Sciences},
  volume    = {372},
  number    = {1711},
  pages     = {20160049},
  year      = {2017},
  doi       = {10.1098/rstb.2016.0049}
}

@book{kanerva1988sparse,
  author    = {Kanerva, Pentti},
  title     = {Sparse Distributed Memory},
  publisher = {MIT Press},
  year      = {1988},
  isbn      = {9780262111324}
}

@inproceedings{wu2018kanerva,
  title     = {The {Kanerva} Machine: A Generative Distributed Memory},
  author    = {Wu, Yan and Wayne, Greg and Graves, Alex and Lillicrap, Timothy},
  booktitle = {International Conference on Learning Representations},
  year      = {2018},
  url       = {https://openreview.net/forum?id=S1HlA-ZAZ}
}

@article{chen2021look,
  title={Where to look: A unified attention model for visual recognition with reinforcement learning},
  author={Chen, Gang},
  journal={arXiv preprint arXiv:2111.07169},
  year={2021}
}

@article{mnih2014recurrent,
  title={Recurrent models of visual attention},
  author={Mnih, Volodymyr and Heess, Nicolas and Graves, Alex and Kavukcuoglu, Koray},
  journal={Advances in neural information processing systems},
  volume={27},
  year={2014}
}

@article{bajcsy2018revisiting,
  title     = {Revisiting active perception},
  author    = {Bajcsy, Ruzena and Aloimonos, Yiannis and Tsotsos, John K.},
  journal   = {Autonomous Robots},
  volume    = {42},
  number    = {2},
  pages     = {177--196},
  year      = {2018},
  doi       = {10.1007/s10514-017-9615-3}
}

@article{schulman2017proximal,
  title={Proximal policy optimization algorithms},
  author={Schulman, John and Wolski, Filip and Dhariwal, Prafulla and Radford, Alec and Klimov, Oleg},
  journal={arXiv preprint arXiv:1707.06347},
  year={2017}
}

@inproceedings{mnih2016asynchronous,
  title={Asynchronous methods for deep reinforcement learning},
  author={Mnih, Volodymyr and Badia, Adria Puigdomenech and Mirza, Mehdi and Graves, Alex and Lillicrap, Timothy and Harley, Tim and Silver, David and Kavukcuoglu, Koray},
  booktitle={International conference on machine learning},
  pages={1928--1937},
  year={2016},
  organization={PmLR}
}

@article{Kirkpatrick2017EWC,
  author    = {Kirkpatrick, James and Pascanu, Razvan and Rabinowitz, Neil and Veness, Joel and Desjardins, Guillaume and Rusu, Andrei A. and Milan, Kieran and Quan, John and Ramalho, Tiago and Grabska-Barwinska, Agnieszka and Hassabis, Demis and Clopath, Claudia and Kumaran, Dharshan and Hadsell, Raia},
  title     = {Overcoming catastrophic forgetting in neural networks},
  journal   = {Proceedings of the National Academy of Sciences},
  volume    = {114},
  number    = {13},
  pages     = {3521-3526},
  year      = {2017},
  doi       = {10.1073/pnas.1611835114},
  URL       = {https://www.pnas.org},
  eprint    = {https://www.pnas.org}
}

@article{fountas2024emllm,
  title     = {Human-inspired Episodic Memory for Infinite Context {LLMs}},
  author    = {Fountas, Zafeirios and Benfeghoul, Martin A. and Oomerjee, Adnan
               and Christopoulou, Fenia and Lampouras, Gerasimos
               and Bou-Ammar, Haitham and Wang, Jun},
  journal   = {arXiv preprint arXiv:2407.09450},
  year      = {2024},
  url       = {https://arxiv.org/abs/2407.09450}
}

@article{zhang2026memrl,
  title     = {{MemRL}: Self-Evolving Agents via Runtime Reinforcement Learning on Episodic Memory},
  author    = {Zhang, Shengtao and Wang, Jiaqian and Zhou, Ruiwen and Liao, Junwei
               and Feng, Yuchen and Li, Zhuo and Zheng, Yujie and Zhang, Weinan and Wen, Ying
               and Li, Zhiyu and Xiong, Feiyu and Qi, Yutao and Tang, Bo and Wen, Muning},
  journal   = {arXiv preprint arXiv:2601.03192},
  year      = {2026},
  url       = {https://arxiv.org/abs/2601.03192}
}

@article{wang2024comprehensive,
  title={A comprehensive survey of continual learning: Theory, method and application},
  author={Wang, Liyuan and Zhang, Xingxing and Su, Hang and Zhu, Jun},
  journal={IEEE transactions on pattern analysis and machine intelligence},
  volume={46},
  number={8},
  pages={5362--5383},
  year={2024},
  publisher={IEEE}
}

@inproceedings{rolnick2019experience,
  title     = {Experience Replay for Continual Learning},
  author    = {Rolnick, David and Ahuja, Arun and Schwarz, Jonathan and
               Lillicrap, Timothy and Wayne, Greg},
  booktitle = {Advances in Neural Information Processing Systems},
  volume    = {32},
  year      = {2019},
  url       = {https://papers.nips.cc/paper_files/paper/2019/hash/fa7cdfad1a5aaf8370ebeda47a1ff1c3-Abstract.html}
}

@inproceedings{bricken2023sparse,
  author    = {Bricken, Trenton and Davies, Xander and Singh, Deepak and Krotov, Dmitry and Kreiman, Gabriel},
  title     = {Sparse Distributed Memory is a Continual Learner},
  booktitle = {International Conference on Learning Representations (ICLR)},
  year      = {2023},
  url       = {https://arxiv.org/abs/2303.11934}
}

@article{shen2023reducing,
  author    = {Shen, Yang and Dasgupta, Sanjoy and Navlakha, Saket},
  title     = {Reducing Catastrophic Forgetting With Associative Learning: A Lesson From Fruit Flies},
  journal   = {Neural Computation},
  volume    = {35},
  number    = {11},
  pages     = {1797--1819},
  year      = {2023},
  note      = {Preprint: arXiv:2107.07617 (2021), ``Algorithmic insights on continual learning from fruit flies''}
}

@article{dohare2021continual,
  title={Continual backprop: Stochastic gradient descent with persistent randomness},
  author={Dohare, Shibhansh and Sutton, Richard S and Mahmood, A Rupam},
  journal={arXiv preprint arXiv:2108.06325},
  year={2021}
}

@article{dasgupta2017neural,
  author    = {Dasgupta, Sanjoy and Stevens, Charles F. and Navlakha, Saket},
  title     = {A neural algorithm for a fundamental computing problem},
  journal   = {Science},
  volume    = {358},
  number    = {6364},
  pages     = {793--796},
  year      = {2017}
}

@inproceedings{janson2022simple,
  author    = {Janson, Paul and Zhang, Wenxuan and Aljundi, Rahaf and Elhoseiny, Mohamed},
  title     = {A Simple Baseline that Questions the Use of Pretrained-Models in Continual Learning},
  booktitle = {NeurIPS 2022 Workshop on Distribution Shifts},
  year      = {2022},
  url       = {https://arxiv.org/abs/2210.04428}
}

@article{zhou2024revisiting,
  author    = {Zhou, Da-Wei and Cai, Zi-Wen and Ye, Han-Jia and Zhan, De-Chuan and Liu, Ziwei},
  title     = {Revisiting Class-Incremental Learning with Pre-Trained Models: Generalizability and Adaptivity are All You Need},
  journal   = {International Journal of Computer Vision},
  year      = {2024},
  note      = {arXiv:2303.07338}
}

@inproceedings{mcdonnell2023ranpac,
  author    = {McDonnell, Mark D. and Gong, Dong and Parvaneh, Amin and Abbasnejad, Ehsan and van den Hengel, Anton},
  title     = {{RanPAC}: Random Projections and Pre-trained Models for Continual Learning},
  booktitle = {Advances in Neural Information Processing Systems (NeurIPS)},
  year      = {2023},
  url       = {https://arxiv.org/abs/2307.02251}
}

\clearpage
\appendix
\section{Hyperparameter optimization}
\label{app:hparams-opt}
To achieve a fair comparison of the methods in their optimal configurations, we performed a limited hyperparameter search of learning rates (\ref{tab:hparams-lr}) and training epochs (\ref{tab:hparams-opt}). Hyperparameters were optimized to the task by training on 400 of the images of each fine-class and evaluating on a validation split of 100 images. The validation splits were not used for final model training.

The decision to allow models to be trained on a larger number of forward passes assumes that the ultimate constraint is the number of samples available rather than the number of learning steps.

The exception is that the RL-STM training epochs were not optimized for the continual setting. 

\begin{table}[htbp]
  \caption{Learning rates explored for each model on Pair A validation split.}
  \label{tab:hparams-lr}
  \centering
  \small
  \begin{tabular}{lp{0.6\linewidth}}
    \toprule
    Model & Learning rates \\
    \midrule
    RL-STM & Not tuned (0.1); learning rate 0.2 with zero momentum tested on validation set for Pair A and found to be equivalent. Few-shot training budget optimized, but not continual. \\
    Sparse supervised STM & 0.1, 0.03, 0.01 \\
    Linear probe & 0.0003, 0.001, 0.003 \\
    NCM & none \\
    FlyModel & Not tuned (0.2) \\
    SDMLP & 0.015, 0.05, 0.15 \\
    LTM-only fine-tuning & 0.01, 0.001, 0.0001, 0.00001 \\
    \bottomrule
  \end{tabular}
\end{table}

\begin{table}[htbp]
  \caption{Training budget optimization on the Pair A validation split. For each model and learning rate, the training epoch budget is selected to optimize mean accuracy on the four fine classes. Bold: chosen. $^\S$Not plateaued at the cap (RL-STM: 48 epochs, four times its starting budget). $^\ddagger$RL-STM at $N = 4$: not plateaued at 24 epochs. --: not evaluated by hyperparameter search rule.}
  \label{tab:hparams-opt}
  \centering
  \scriptsize
  \setlength{\tabcolsep}{0.7pt}
  \begin{tabular}{llrrrrrr}
    \toprule
    & & & \multicolumn{5}{c}{Few-shot, $N =$} \\
    \cmidrule(lr){4-8}
    Model & LR & Continual & 1 & 4 & 16 & 64 & 400 \\
    \midrule
    RL-STM & 0.1 & 12 (fixed) & \textbf{1 (0.571)} & \textbf{24 (0.646)$^\ddagger$} & \textbf{26 (0.721)$^\S$} & \textbf{21 (0.735)$^\S$} & \textbf{21 (0.731)} \\
    Sparse supervised STM & 0.01 & 12 (fixed) & \textbf{1 (0.714)} & \textbf{8 (0.721)} & \textbf{42 (0.736)} & \textbf{6 (0.735)} & \textbf{1 (0.740)} \\
    FlyModel & -- & \textbf{1 (0.749)} & \textbf{1 (0.745)} & \textbf{1 (0.746)} & \textbf{1 (0.744)} & \textbf{3 (0.756)} & \textbf{1 (0.770)} \\
    Linear probe & 0.0003 & 24 (0.770) & 1 (0.659) & 2060 (0.705) & 471 (0.721) & 126 (0.748) & 24 (0.755) \\
    Linear probe & 0.001 & \textbf{8 (0.775)} & \textbf{1 (0.658)} & \textbf{589 (0.705)} & \textbf{148 (0.722)} & \textbf{36 (0.745)} & \textbf{8 (0.759)} \\
    Linear probe & 0.003 & 3 (0.776) & 1 (0.658) & 185 (0.705) & 45 (0.720) & 12 (0.747) & 3 (0.758) \\
    LTM-only fine-tuning & 0.0001 & \textbf{10 (0.796)} & 1 (0.401) & 39802 (0.554)$^\S$ & \textbf{7597 (0.673)} & \textbf{236 (0.703)} & \textbf{13 (0.749)} \\
    LTM-only fine-tuning & 0.001 & 14 (0.773) & \textbf{25246 (0.479)} & \textbf{24 (0.601)} & 38 (0.650) & 9 (0.678) & 4 (0.698) \\
    LTM-only fine-tuning & 0.01 & -- & 4 (0.505) & 65 (0.522) & -- & -- & -- \\
    LTM-only fine-tuning & 0.00001 & -- & -- & -- & -- & -- & 120 (0.749) \\
    NCM & -- & \textbf{49 (0.781)} & \textbf{1 (0.742)} & \textbf{702 (0.756)} & \textbf{521 (0.765)} & \textbf{77 (0.766)} & \textbf{17 (0.770)} \\
    SDMLP & 0.015 & 51 (0.772) & 1 (0.744) & 2551 (0.756) & 1 (0.743) & 190 (0.760) & 40 (0.761) \\
    SDMLP & 0.05 & \textbf{16 (0.773)} & \textbf{1 (0.744)} & \textbf{702 (0.752)} & \textbf{1 (0.743)} & \textbf{58 (0.760)} & \textbf{14 (0.767)} \\
    SDMLP & 0.15 & 6 (0.774) & 1 (0.744) & 235 (0.752) & 45 (0.745) & 20 (0.762) & 6 (0.769) \\
    \bottomrule
  \end{tabular}
\end{table}

\clearpage

\section{Details of the LTM model}
\label{app:ltm}
The LTM is a ResNet-18 model \citep{he2016deep} adapted to CIFAR image sizes. The architecture is described in table \ref{tab:ltm-layers}. Input images are scaled to unit range by dividing by 255; no mean/standard-deviation normalization and no data augmentation are used at any stage, for any model.

Table \ref{tab:ltm} describes hyperparameters used in LTM pre-training.

\begin{table}[H]
  \caption{Long Term Memory configuration. Note there are 2 residual blocks per stage.}
  \label{tab:ltm-layers}
  \centering
  \begin{tabular}{lll}
    \toprule
    Component & Output dimension      \\
    \midrule
    Input & $3\times32\times32$ \\
    Stem: $3\times3$ conv., 64 channels, stride 1 & $64\times32\times32$ \\
    Stage 1: $3\times3$ conv., 64 channels, stride 1& $64\times32\times32$ \\
    Stage 2: $3\times3$ conv., 128 channels, stride 2 & $128\times16\times16$ \\
    Multiplicative ``bias'' applied here. & $128\times16\times16$ \\
    Stage 3: $3\times3$ conv., 256 channels, stride 2 & $256\times8\times8$ \\
    Stage 4: $3\times3$ conv., 512 channels, stride 2 & $512\times4\times4$ \\
    Mean pooling & $512\times1\times1$  (``encoding'') \\
    Linear (``classifier'') & 20 classes \\
    \bottomrule
  \end{tabular}
\end{table}

\begin{table}[H]
  \caption{Long Term Memory hyperparameters}
  \label{tab:ltm}
  \centering
  \begin{tabular}{ll}
    \toprule
    Name     & Value      \\
    \midrule
    Pretraining mini-batch size & 128 (16 in fine-tuning) \\
    Optimizer & Adam \\
    Learning rate & 0.001 (varied for fine-tuning) \\
    \bottomrule
  \end{tabular}
\end{table}

\clearpage

\section{RL-STM Hyperparameters}
\label{app:rl-stm-param}
Table \ref{tab:stm} contains RL-STM hyperparameters used in all experiments. A learning rate of 0.2 with momentum 0 was found to yield similar performance in ad-hoc testing (no advantage or disadvantage, contrary to findings in \citep{bricken2023sparse}).

Note: Actor loss scaling is crucial to provide a sufficiently stable policy for Critic learning. No policy entropy management was implemented.

\begin{table}[H]
  \caption{RL-STM hyperparameters}
  \label{tab:stm}
  \centering
  \begin{tabular}{ll}
    \toprule
    Name     & Value      \\
    \midrule
    Pretraining mini-batch size & 16 \\
    Fine-tuning mini-batch size & 16 \\
    Optimizer & SGD \\
    Learning rate & 0.1 \\
    Momentum & 0.5 \\
    Memory capacity $h$ & 1000 \\
    Activation sparsity $k$ & 32 \\
    Episode length (steps) $n$ & 8 \\
    Context window length & 8 \\
    Discount factor & 0.5 \\
    Continuous policy Std. Dev. & 0.5 \\
    Actor loss scale & 0.01 \\
    Critic loss scale & 1.0 \\
    Entropy loss & None \\
    \bottomrule
  \end{tabular}
\end{table}

\clearpage

\section{Details of other STM variants}
\label{app:other-stm}

All STM variants consume the 512-dimensional stage-4 pooled encoding of the frozen LTM, computed with the LTM in evaluation mode. Like the RL-STM and LTM-only models, all STM variants produce 20-way classification logits. 

\subsection{FlyModel}
See \citep{shen2023reducing} for details. Each encoding is min-max scaled, projected by a fixed binary random matrix onto 20,000 units (64 non-zero inputs per unit), and thresholded to the top 1\% of units. Output weights are updated by a Hebbian rule: the active units are added to the row of the target class with learning rate 0.2 and clipped at 1. No gradients are used.

\subsection{SDMLP}
See \citep{bricken2023sparse} for details. A hidden layer of 1000 units without bias, with non-negative weights and L2-normalized addresses and inputs, followed by subtractive Top-K inhibition with $k=32$ (the same $k$ as the RL and sparse supervised STM, and used in the cited work) and the ``GABA switch'', then a linear read-out. It is trained by stochastic gradient descent on a cross-entropy loss (learning rate 0.05, momentum 0.9). The GABA switch threshold is scaled to the size of our pre-training set (12,000 activations per unit) since the original value assumes a much larger dataset.

\subsection{Linear probe and NCM}
The linear probe is a single linear layer initialized from the LTM's own classifier and trained by SGD with a cross-entropy loss. The NCM classifier maintains a running mean encoding per coarse class and predicts by cosine similarity; classes not yet seen are excluded.

\end{document}